\documentclass[10pt, a4paper]{article}
\usepackage{lrec2022} 
\usepackage{multibib}
\newcites{languageresource}{Language Resources}
\usepackage{graphicx}
\usepackage{tabularx}
\usepackage{soul}
\usepackage{titlesec}
\titleformat{\section}{\normalfont\large\bfseries\center}{\thesection.}{1em}{}
\titleformat{\subsection}{\normalfont\SmallTitleFont\bfseries\raggedright}{\thesubsection.}{1em}{}
\titleformat{\subsubsection}{\normalfont\normalsize\bfseries\raggedright}{\thesubsubsection.}{1em}{}
\renewcommand\thesection{\arabic{section}}
\renewcommand\thesubsection{\thesection.\arabic{subsection}}
\renewcommand\thesubsubsection{\thesubsection.\arabic{subsubsection}}

\usepackage{epstopdf}
\usepackage[utf8]{inputenc}

\usepackage{hyperref}
\usepackage{xstring}

\usepackage{color}
\usepackage{pgfplots}

\title{BERTifying Sinhala - A Comprehensive Analysis of Pre-trained Language Models for Sinhala Text Classification \\ }

\name{Vinura Dhananjaya, Piyumal Demotte, Surangika Ranathunga, Sanath Jayasena} 

\address{Department of Computer Science and Engineering \\
        University of Moratuwa, Katubedda 10400, Sri Lanka \\
         dhananjayagv.21@uom.lk, piyumalanthony.16@cse.mrt.ac.lk, surangika@cse.mrt.ac.lk, sanath@uom.lk\\}

\abstract{
This research provides the first comprehensive analysis of the performance of pre-trained language models for Sinhala text classification. We test on a set of different Sinhala text classification tasks  and our analysis shows that out of the pre-trained multilingual models that include Sinhala (XLM-R, LaBSE, and LASER), XLM-R is the best model by far for Sinhala text classification. We also pre-train two RoBERTa-based monolingual Sinhala models, which are far superior to the existing pre-trained language models for Sinhala. We show that when fine-tuned, these pre-trained language models set a very strong baseline for Sinhala text classification and are robust in situations where labeled data is insufficient for fine-tuning. We further provide a set of recommendations for using pre-trained models for Sinhala text classification. We also introduce new annotated datasets useful for future research in Sinhala text classification and publicly release our pre-trained models. 
 \\ \newline \Keywords{Language Models, Monolingual, Multilingual, Text Classification} }

\begin{document}

\maketitleabstract

\section{Introduction}
Large-scale monolingual pre-trained language models (MonoLMs) such as BERT~~\cite{BERT}, RoBERTa~\cite{roberta}, and their multilingual descendants mBERT~\cite{BERT} and XLM-R~\cite{xlmr} ( respectively), have shown promising results for high-resource as well as low-resource languages, particularly for text classification~\cite{wudredze-beto,aguilar}. Following this early success, many empirical studies have been carried out to determine the performance of these models on different settings. However, most of these studies focused on a limited number of languages. Moreover, experimental results show that the performance of the pre-trained Multilingual Language Models (MultiLMs) depends on many factors - such as the amount of language data used during the pre-training, the relatedness of a language to the other languages in the pre-trained model, language characteristics, the typography of the language, and the amount of fine-tuning data used~\cite{wudredze,primer}. Thus, the results reported in these studies cannot be generalized across languages. For MonoLMs, a major deciding factor is the amount of monolingual data used in the model pre-training stage~\cite{pfeiffer}.\\
As noted by~\newcite{soriaquochi}, `a Natural Language Processing (NLP) system can be measured only in terms of its usefulness for the end-users'. In other words, the usefulness of pre-trained language models on a language depends on their ability to provide acceptable results for the NLP tasks of the considered language, but not by their performance on some other set of languages. Thus, it is imperative that we carry out extensive evaluation of the pre-trained models for the specific languages of interest.

Sinhala is an Indo-Aryan language primarily used by a population of about 20 million, in the small island nation of Sri Lanka. According to \newcite{joshi}'s language categorization, Sinhala has been given class 1, meaning an extremely low-resource language. This is not surprising - not only the available language resources, but also the amount of research is scarce for Sinhala~\cite{nisansa-survey}. However, Sinhala has been fortunate to get included in pre-trained MultiLMs such as XLM-R, LASER~\cite{laser} LaBSE~\cite{labse}, mT5~\cite{mt5} and mBART~\cite{mbart}\footnote{Sinhala is not included in mBERT}. There also exist monolingual Sinhala pre-trained models\footnote{SinBerto; https://huggingface.co/Kalindu/SinBerto and SinhalaBERTo; https://huggingface.co/keshan/SinhalaBERTo}. However, they have been pre-trained on relatively small Sinhala corpora. No results of using these monolingual or MultiLMs for Sinhala text classification have been reported so far. Thus, the effectiveness of these models for Sinhala text classification is not known yet.

In this research, we build two RoBERTa based pre-trained language models. Compared to the  existing Sinhala pre-trained models, our models are trained with a much larger corpus\footnote{https://github.com/brainsharks-fyp17/sinhala-dataset-creation, corpus at; https://bit.ly/3OBVuoU}. Our objective is to identify the best pre-trained model for different Sinhala text classification tasks. Thus, the built Sinhala RoBERTa models are compared against the MultiLMs that include Sinhala; LASER, XLM-R, and LaBSE.\\
We use 4 different classification tasks, namely, sentiment analysis with a 4-class sentiment dataset ~\cite{demottesent}, news category classification with a 5-class dataset~\cite{nisansa}, a 9-class news source classification and a 4-class writing style classification task. Dataset for the last two tasks have been prepared by us.\\
Based on our initial experiment results, we identify XLM-R as the best MultiLM for Sinhala text classification. From the experiments with our MonoLMs and XLM-R, we observe that the XLM-R-large model yields consistently better results than our MonoLMs. We further observe that our MonoLMs perform better than XLM-R-base when the fine-tuning dataset size is small. Based on our results, we provide a set of recommendations, which would be useful for future research on Sinhala text classification. Moreover, we set new baselines for all the selected Sinhala text classification tasks. 

We publicly release the trained Sinhala RoBERTa models (which are referred to as SinBERT-large and SinBERT-small, from here onwards) via Hugginface\footnote{SinBERT-small-https://huggingface.co/NLPC-UOM/SinBERT-small}, \footnote{SinBERT-large-https://huggingface.co/NLPC-UOM/SinBERT-large}. The annotated datasets for Sinhala news source classification\footnote{https://huggingface.co/datasets/NLPC-UOM/Sinhala-News-Source-classification}, news category classification\footnote{https://huggingface.co/datasets/NLPC-UOM/Sinhala-News-Category-classification} and writing style classification\footnote{https://huggingface.co/datasets/NLPC-UOM/Writing-style-classification} are also publicly released.

\section{Pre-trained Language Models}

Pre-trained language models aim at exploiting large unlabeled corpora to learn text representations at scale, such that the trained models can be fine-tuned on relatively smaller, labeled datasets for downstream tasks. LASER and ELMo~\cite{elmo} were amongst the initial pre-trained language models based on neural network architectures capable of learning long-term dependencies in sequences, such as GRU~\cite{gru} and Bi-LSTM~\cite{bilstm}. \\
The inception of the Transformer~\cite{transformer} architecture propelled the creation of state-of-the-art language models. BERT was pre-trained with Masked Language Modelling (MLM) and Next Sentence Prediction (NSP) tasks on the large BookCorpus dataset~\cite{bookcorpus} and English Wikipedia corpus. It became a basis for many other language models that followed up. RoBERTa, which stands for Robustly Optimized BERT Pre-training Approach, is a Transformer architecture similar to BERT. The first RoBERTa model trained for the English language introduced modifications over the pre-training process used in BERT. These include the use of larger batch sizes, removal of the NSP pre-training objective, using longer sequences for pre-training, training the model for a longer period and using a dynamic masking pattern for the MLM task.\\
Multilingual BERT (mBERT) was released as the multilingual variant of BERT pre-trained on corpora from 104 languages. LaBSE is another MultiLM, which uses a dual-encoder architecture based on BERT and supporting 109 languages. It has been trained on MLM and Translation Language Model (TLM) objectives. XLM-R is a MultiLM pre-trained on CommonCrawl~\cite{ccdata} based on XLM~\cite{xlm}, and supports 100 languages. It uses MLM as its pre-training task.\\
While models such as mBERT and XLM-R are encoder-only models, T5~\cite{t5} and BART~\cite{bart} (and their multilingual variants mT5 and mBART) contain an encoder and a decoder, and are ideal for sequence-to-sequence tasks such as text summarization. However, their usage in text classification tasks is comparatively less.

\section{Related Work}
\subsection{Text Classification with Pre-trained Models}
Following the success of pre-trained models for English, similar models have been built for some other languages. Some examples are, FlauBERT (French)~\cite{flaubert}, FinBERT (Finnish)~\cite{finbert}, AraBERT (Arabic)~\cite{arabert}, PhoBERT (Vietnamese)~\cite{phobert} and AfriBERT (Afrikaans)~\cite{afrikaans}. Each model has been able to set a new state-of-the-art for a variety of NLP tasks for the corresponding language. Some have compared the MonoLMs they built with the MultiLMs~\cite{phobert,flaubert,finbert,uralic}. However, it cannot be concluded that the MonoLMs are better than MultiLMs, and vice-versa, across different tasks and languages.
\newcite{wudredze} attributed this discrepancy solely to the amount of data used to pre-train the models. However, \newcite{pfeiffer}'s findings suggest that the pre-trained tokenizer also plays an important role in downstream task performance, as well as the selected tasks.

In addition to the above research that compared monolingual and MultiLMs, there is a plethora of research that analysed various aspects of pre-trained MultiLMs across various NLP tasks and languages. \newcite{aguilar} and \newcite{lauscher} showed that the performance of the multilingual pre-trained models is not consistent across the NLP tasks. According to \newcite{aguilar}, these models are better at syntactic analysis as opposed to semantic analysis. \newcite{groenwold} and \newcite{lauscher} showed that the performance of a pre-trained model on a given language is heavily influenced by the language family. In other words, more related languages are included in the model is beneficial for a language. As a result, pre-trained models have been shown to perform better for Indo-European languages~\cite{xtreme}. Some others experimented on different conditions such as zero shot performance on languages that are  included in the pre-trained model~\cite{xtreme,wudredze-beto,embrahimi-americasnli,litschko}, and performance on languages not included in the pre-trained models~\cite{ebrahimikann}.\\
Although pre-trained MultiLMs such as mBERT and XLM-R are a very attractive option for low-resource language computing, they have bounded capacity with respect to the number of languages that can be included in the model. This is commonly known as the ``curse of multilinguality"~\cite{xlmr}. Moreover, low resource languages are mostly underrepresented in MultiLMs (i.e. the pre-trained models include comparatively low amounts of training data from these languages), which makes these models to under-perform for those languages compared to high resource languages included in MultiLMs~\cite{wudredze}. The alternative is to train MultiLMs only for a set of related languages. IndicBERT~\cite{indicbert}\footnote{https://indicnlp.ai4bharat.org/indic-bert/} is a very good example for this. When the average result for a particular task across the indic languages is considered, IndicBERT outperforms both mBERT and XLM-R by a substantial margin in tasks such as question answering and cross-lingual sentence retrieval.

\subsection{Sinhala Text Classification}
Being a fusional language and having rich linguistic features, the Sinhala language inherits a certain complexity of language understanding added to its scarcity of resources. Research in Sinhala text classification has been mainly limited to traditional approaches. Experiments with  Machine Learning methods such as Support Vector Machines (SVM) were carried out by~\cite{gallage}. Furthermore, approaches such as rule-based systems~\cite{lakmali}, a stop word extraction method for text classification using TF-IDF~\cite{haddela}, Feed-forward Neural network based system~\cite{medagoda} and a Word2Vec based approach\footnote{http://bit.ly/2QKI9Np} have also been followed. \newcite{chathuranga} proposed a method for Sinhala text classification based on a lexicon. \newcite{liyanage} are the first to experiment with Deep Learning techniques such as LSTM networks as well as Convolutional Neural Networks (CNN) based methods for Sinhala sentiment classification. \newcite{demotte} also proposed a LSTM-based system for Sinhala text classification based on S-LSTMs~\cite{slstm}. More recently, \newcite{demottesent} empirically analysed RNN, Bi-LSTM and Capsule Networks for Sinhala news text sentiment classification. SinBERTo and Sinhala-RoBERTa are two separately pre-trained RoBERTa based MonoLMs for Sinhala, which have been released recently. They do not have related work published, nor have been used in text classification, to the best of our knowledge. Although encoder-based pre-trained models have not been used for Sinhala, mBART has shown exceptionally good results for Machine Translation that involves Sinhala~\cite{sarubi}.



\section{SinBERT Model}

\subsection{Pre-training  and Fine-tuning Setup}
\label{pretrainingsetup}

RoBERTa has shown improved results over other competitive models for the GLUE benchmark~\cite{glue}, specifically for classification tasks. Hence, we build our Sinhala MonoLMs based on RoBERTa. We use Huggingface's\footnote{https://huggingface.co/}  Transformers libraries in Pytorch~\cite{pytorch} to pre-train our RoBERTa models\footnote{We publicly release the pre-training and fine-tuning codes on https://github.com/nlpcuom/Sinhala-text-classification}. We use AdamW~\cite{adamw} as the optimizer with a learning rate of 1e-4, a batch size of 16 and a maximum of 2 training epochs to pre-train the models. We introduce two variants of our model; SinBERT-small containing 6 hidden layers and SinBERT-large containing 12 hidden layers. Parameters of the two models are shown in Table~\ref{statmodels}.

Fine-tuning hyper-parameters are given in Table~\ref{stathyper}. We used the standard fine-tuning process, where the [CLS] token output from the pre-trained model's encoder was fed to a feed-forward neural network based classifier. For Sinhala monolingual models, we use Huggingface's default classifier for RoBERTa models which consists of a linear layer, a dropout layer preceded by the pooled output from the model encoder layer. A linear layer preceded by a dropout layer was used as the classifier head for LASER and LaBSE. For XLM-R-large, we use a batch size of 8 due to hardware constraints. \\
We report the macro-averaged F1-score over 5 different randomly-initialized runs for each experiment using 4:1 train/test splits of the datasets. For LaBSE and LASER we use only 3 randomly-initialized runs as their performance is well below to that of XLM-R and the monolingual models.
All the pre-training and fine-tuning were conducted on a single Nvidia Quadro RTX 6000 (24GB) GPU.

\subsection{Sinhala Corpus used for SinBERT pre-training}

SinBERT models are pre-trained using ``sin-cc-15M" corpus\footnote{https://tinyurl.com/42un7a9y}. At present, it is the largest Sinhala monolingual corpus available to the best of our knowledge. The dataset comprises of 15.7 million sentences extracted from 3 sources: CC-100, OSCAR and raw text data from Sinhala news web sites. CC-100 dataset contains 3.7GB of data for Sinhala and OSCAR contains 802MB of Sinhala text including duplicated text. The raw news data extracted from Sinhala news sites is 413MB in size. The final sin-cc-15M dataset has been cleaned of other language words/characters and invalid characters. Cleaned dataset statistics are shown in Table~\ref{statcorpus}.

\begin{table}[!h]
\begin{center}
\begin{tabularx}{\columnwidth}{|l|X|X|}

      \hline
      Number of words & 192.6M\\
      \hline
      Number of unique words & 2.7M\\
       \hline
     Number of sentences & 15.7M\\
      \hline
     Average number of words/sentence & 12.2\\
      \hline

\end{tabularx}
\caption{Statistics of the pre-training corpus}
\label{statcorpus}
 \end{center}
\end{table}

\begin{table}[!h]
\begin{center}
\begin{tabularx}{\columnwidth}{|l|X|X|}

      \hline
      &SinBERT-small&SinBERT-large\\
      \hline
      Hidden layers & 6 & 12\\
      \hline
      Attention heads & 6 & 12\\
       \hline
      Max. Position embeddings & 514 & 514\\
      \hline
     Vocabulary size & 30000 & 52000\\
      \hline
     Number of Parameters & 66.5M & 125.9M\\

      \hline

\end{tabularx}
\caption{Parameters of two SinBERT models}
\label{statmodels}
 \end{center}
\end{table}

\section{Experiments}

\subsection{Model Selection}
We compare the trained RoBERTa models with three MultiLMs: XLM-R-base and large, LaBSE\footnote{https://tfhub.dev/google/LaBSE/2}, and LASER\footnote{https://github.com/facebookresearch/LASER}. For other Sinhala MonoLMs, we take two RoBERTa based models publicly available in Huggingface; SinBERTo and SinhalaBERTo. Both have a vocabulary size of 52 000 and a similar model architecture (6 hidden layers, 12 attention heads, max. position embedding size of 514). However, SinBerto has been trained on a small news corpus while SinhalaBERTo has been trained on a much larger deduplicated Sinhala OSCAR dataset. There are two other Sinhala MonoLMs available in Huggingface (sinhala-Roberta-Oscar\footnote{https://huggingface.co/keshan/sinhala-roberta-oscar} and sinhala-roberta-mc4\footnote{https://huggingface.co/keshan/sinhala-roberta-mc4}), however, their vocabulary sizes are smaller.

\subsection{Fine-tuning Tasks}
We use four sentence/document level classification tasks. For the first two tasks given below, annotated data was already available. For the other two tasks, we prepared the annotated data from the raw corpora.

\subsubsection{Sentiment Analysis}
\label{sc_sa}
We use the sentiment dataset published by~\newcite{demottesent} for the sentiment classification task. This dataset consists of user comments published in response to online news articles. Each user comment is labeled using four classes (\textit{positive, negative, neutral, conflict}). Thus this can be considered as a document classification task. This is an extension to the dataset introduced by~\newcite{liyanage}, and carries a Cohen's Kappa value of 0.65. \newcite{demottesent} reported a baseline for this task using RNNs and capsule networks.
\subsubsection{News Category Classification}
\label{sc_nc}
The news category dataset\footnote{https://osf.io/tdb84/} contains sentences extracted from 5 different categories of news (\textit{Business, Political, Entertainment, Science and Technology, Sports}) with 1019 maximum number of sentences and 438 minimum sentences for a class~\cite{nisansa}. Thus this is a sentence classification task. The publicly available version of the news category dataset has not been processed. Hence, we pre-process the dataset and  remove sentences that contain English only words and sentences having a length less than 3 words (e.g.- Names of places, celebrities).~\newcite{nisansa} reported an accuracy score result as a baseline for this task using an approach based on SAFS3 algorithm~\cite{safs3}.

\subsubsection{Writing Style Classification}
\label{sc_ws}
We extracted text from \newcite{nisansa2}'s large Sinhala corpus\footnote{https://osf.io/a5quv/files/; publicly available files only contain a portion of the corpora described in their paper}, which contains text spanning across a set of genres. For writing style classification, we select text belonging to 4 categories (\textit{News, Academic, Blog, Creative}). This is a document classification task. We process the extracted text by deduplicating, removing English only text and very long text (length larger than 3500 characters). Since the dataset contains long text, we use truncation to fit them into the models. No evaluation has been presented for this dataset.\\

\subsubsection{News Source Classification}
\label{sc_ns}
This is an annotated dataset newly compiled by us. The news source dataset comprises news headlines in Sinhala, scraped from 9 different Sinhala news web sites (Sri Lanka Army\footnote{https://www.army.lk/}, Dinamina\footnote{http://www.dinamina.lk/}, GossipLanka\footnote{https://www.gossiplankanews.com/}, Hiru\footnote{https://www.hirunews.lk/}, ITN\footnote{https://www.itnnews.lk/}, Lankapuwath\footnote{http://sinhala.lankapuvath.lk/}, NewsLK\footnote{https://www.news.lk/}, Newsfirst\footnote{https://www.newsfirst.lk/}, World Socialist Web Site-Sinhala\footnote{https://www.wsws.org/si})  on the Internet. We reduce the amount of data in the original web-scraped news-source dataset~\cite{newscorpus} in order to handle the class imbalance. We also remove one news source (Sinhala Wikipedia) from the originally scraped dataset as it mostly contains invalid characters, numbers and single word sentences. This is a sentence classification task (since we classify the news headings).



\begin{table}[!h]
\begin{center}
\begin{tabularx}{\columnwidth}{|l|X|X|X|}
      \hline
      Parameter&SinBER T&XLM-R&Other\\
      \hline
      Starting learning rate&1e-5&5e-6&5e-5, 1e-5\\
      \hline
      Batch size&16&16, 8&16\\
      \hline
      No. of epochs&10&5&5\\
      \hline
      Optimizer&AdamW&AdamW&AdamW\\
      \hline
\end{tabularx}
\caption{Hyperparameters for model fine-tuning}
\label{stathyper}
 \end{center}
\end{table}

\begin{table}[!h]
\begin{center}
\begin{tabularx}{\columnwidth}{|l|X|X|X|}

      \hline
      Dataset&No.of data points&No. of classes&Average text length\\
      \hline
      Sentiment&15059&4&21.66\\
      \hline
      News sources &24093&9&8.42\\
      \hline
     News categories &3327&5&23.49\\
      \hline
      Writing style&12514&4&181.97\\
      \hline

\end{tabularx}
\caption{Statistics of the Fine-tuning datasets used}
 \end{center}
\end{table}

\begin{table}[!h]
\begin{center}
\begin{tabularx}{\columnwidth}{|l|X|X|X|}

      \hline
      Dataset&Maximum data points&Minimum data points\\
      \hline
      Sentiment&7665&1911\\
      \hline
      News sources &3109&1541\\
      \hline
     News categories &1019&438\\
      \hline
      Writing style&4463&2111\\
      \hline

\end{tabularx}
\caption{Statistics of the Fine-tuning datasets used-max/min data points in a class}
 \end{center}
\end{table}

\subsection{Evaluation}

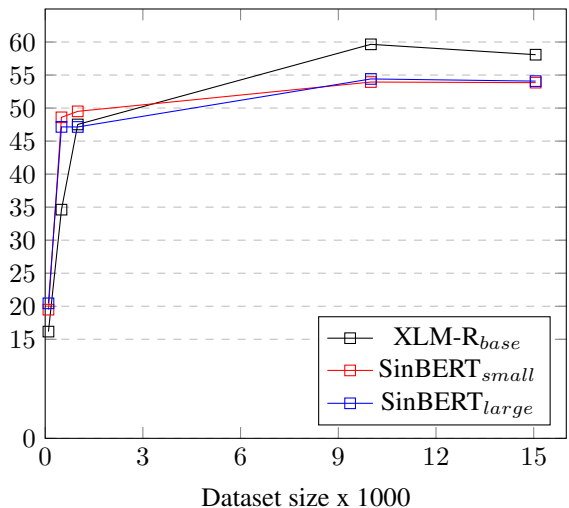
\begin{figure}
\begin{tikzpicture}
\begin{axis}[
    xlabel={Dataset size x 1000},
    xmin=0, xmax=16,
    ymin=0, ymax=65,
    xtick={0,3,6,9,12,15,18},
    ytick={0,15,20,25,30,35,40,45,50,55,60},
    legend pos=south east,
    ymajorgrids=true,
    grid style=dashed,
]

\addplot[
    color=black,
    mark=square,
    ]
    coordinates {
    (0.1,16.16)(0.5,34.62)(1,47.52)(10,59.65)(15.059,58.08)
    };
    \addlegendentry{XLM-R$_{base}$}
    
\addplot[
    color=red,
    mark=square,
    ]
    coordinates {
    (0.1,19.48)(0.500,48.6)(1,49.504)(10,53.94)(15.059,53.85)
    };
    \addlegendentry{SinBERT$_{small}$}
    
\addplot[
    color=blue,
    mark=square,
    ]
    coordinates {
    (0.100,20.44)(0.500,47.15)(1,47.14)(10,54.41)(15.059,54.08)
    };
    \addlegendentry{SinBERT$_{large}$}
    
\end{axis}
\end{tikzpicture}
\caption{Results in macro-F1 score for varying dataset sizes in sentiment classification task with SinBERT models and XLM-R-base}
\label{fig.1}
\end{figure}

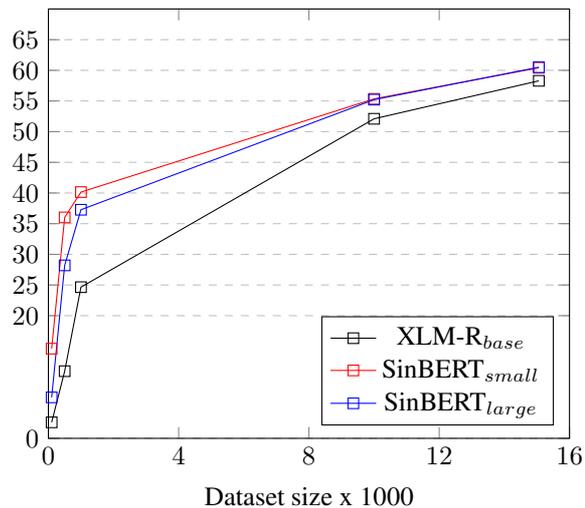
\begin{figure}
\begin{tikzpicture}
\begin{axis}[
    xlabel={Dataset size x 1000},
    xmin=0, xmax=16,
    ymin=0, ymax=70,
    xtick={0,4,8,12,16,20,24,25},
    ytick={0,20,25,30,35,40,45,50,55,60,65},
    legend pos=south east,
    ymajorgrids=true,
    grid style=dashed,
]

\addplot[
    color=black,
    mark=square,
    ]
    coordinates {
    (0.1,2.61)(0.5,10.95)(1,24.65)(10,52.1)(15.059,58.29)
    };
    \addlegendentry{XLM-R$_{base}$}
    
\addplot[
    color=red,
    mark=square,
    ]
    coordinates {
    (0.1,14.62)(0.500,36.02)(1,40.15)(10,55.35)(15.059,60.42)
    };
    \addlegendentry{SinBERT$_{small}$}
    
\addplot[
    color=blue,
    mark=square,
    ]
    coordinates {
    (0.100,6.68)(0.500,28.2)(1,37.27)(10,55.22)(15.059,60.51)
    };
    \addlegendentry{SinBERT$_{large}$}
    
\end{axis}
\end{tikzpicture}
\caption{Results in macro-F1 score for varying dataset sizes in news source classification task with SinBERT models and XLM-R-base}
\label{fig.2}
\end{figure}

Table \ref{resultmain} reports the results for each of our tasks performed using the selected models. We also report the current baseline results for each of the tasks, whenever available. Note that the baseline results for sentiment analysis has been reported with weighted-F1. For a meaningful comparison, we report the same results in the metric used in the baseline paper as well. Since the largest selected model (XLM-R-large) demands high levels of GPU resources to run on, we limit XLM-R-large to the experiments reported in Table~\ref{resultmain}.\\
Results of LaBSE and LASER are consistently lower than both our MonoLMs, as well as the XLM-R models across all the tasks. In fact, LaBSE has a very poor performance across all the tasks. Thus, we can safely advise against using these models for Sinhala text classification. XLM-R-large outperforms the base version in all the tasks, which is not surprising. However, this margin is small in tasks such as sentiment analysis and news category classification.\\
The SinBERT models outperform the existing Sinhala pre-trained models, thus establishing our models as the best monolingual pre-trained models for Sinhala text classification. Interestingly, our large model has only a very small gain against the small model. We believe this is due to the small size of the Sinhala corpus used to pre-train the models- the dataset is not sufficient to properly train the large model. Considering the low performance gains and the time and memory complexity of fine-tuning the SinBERT-large model, we advise the use of the small model in future Sinhala text classification tasks.\\
It can be seen that the XLM-R-large model outperforms both of our SinBERT models. Thus, if the hardware requirements (see Section \ref{pretrainingsetup}) can be satisfied, the best model choice for Sinhala text classification is the XLM-R-large model. However, in a constrained hardware setting, either the XLM-R-base model or the SinBERT-small model can be used. Specifically, XLM-R-base model outperforms the SinBERT-small for all the tasks except the news source categorisation task. We believe this is because the raw news source dataset was included in the SinBERT model training. This is also an important finding. Even if the annotated data amount is small, if the corresponding raw corpus can be included while model pre-training, a result increase can be expected.\\
In the XLM-R models, Sinhala data attributes to only $\sim$0.15\% of the total pre-trained corpora. Moreover, Sinhala has its own script and characteristics. Compared to this low representation and the uniqueness of the language, XLM-R performance on Sinhala is impressive. Sinhala is an Indo-Aryan language and the model contains a relatively higher proportion of data from related languages such as Hindi, and an even higher proportion of distantly related Indo-European languages. This might have contributed to the high performance gains for Sinhala.\\
Figures \ref{fig.1} - \ref{fig.4} depict the macro-F1 score for XLM-R-base and SinBERT models with varying dataset sizes. We vary the dataset sizes as 100, 500, 1000, 10000 and total dataset size (for the news type categorization experiment, the experiment with dataset size of 10000 is skipped since its total dataset size is below 10000). All the graphs show that for smaller dataset sizes, XLM-R-base model lags behind SinBERT models but catches up quickly as the dataset size increases. Thus, if the annotated dataset is extremely small, using the SinBERT-small model would be more fruitful. \\
Even the XLM-R-base model and the SinBERT-small model outperform the current baselines for sentiment analysis. Finally, text classification with the XLM-R-large results establish a new (strong) baseline for each of the considered tasks.\\ 
Out of the four contrasting classification tasks and datasets that were used, sentiment analysis and news source classification task yield the lowest F1-scores, thus they can be considered as the most difficult tasks. News source prediction is a difficult task for humans as well unless the news headlines carry distinguished styles of writing or keywords in them. In our dataset, Army news website headlines are comparatively shorter in length and contains a small set of frequently used words, which makes it easier to be identified by the model. The sentiment analysis dataset contains one under-represented class label \textit{conflict}, which makes it more challenging for the model to differentiate between the sentiment classes. In the news categories and writing style datasets, the sentences/documents in both datasets contain distinct sets of words or keywords, which makes it easier for the model to predict the classes. \\

\begin{table*}[!h]
\begin{center}
\small
\begin{tabularx}{1.75\columnwidth}{|l|X|X|X|X|}
    \hline
    Model&Sentiment&News sources&News categories&Writing style\\
      \hline
      \textit{Baseline}&59.42$_{w.F1}$&-&-&-\\
      \hline
      LaBSE&20.63&11.85&24.09&-\\
      \hline
     LASER &54.07&28.84&48.54&87.06\\
      \hline
       XLM-R$_{base}$&58.08&58.29&85.12&96.89\\
      \hline
       XLM-R$_{large}$&\textbf{60.45}~\small{(68.1$_{w.F1}$)}&\textbf{61.84}&\textbf{89.54}&\textbf{98.41}\\
      \hline
       SinBERTo&50.83&57.22&78.07&93.84\\
      \hline
       SinhalaBERTo&49.71&57.34&82.73&94.10\\
      \hline
       SinBERT$_{small}$&53.85&60.42&84.75&95.00\\
      \hline
       SinBERT$_{large}$&54.08&60.51&85.19&95.49\\
      \hline

\end{tabularx}
\caption{macro-F1 scores for the selected models on 4 classification tasks.}
\label{resultmain}
 \end{center}
\end{table*}






%

\begin{figure}
\begin{tikzpicture}
\begin{axis}[
    xlabel={Dataset size x 1000},
    xmin=0, xmax=4,
    ymin=0, ymax=100,
    xtick={0,1,2,3},
    ytick={0,20,30,40,50,60,70,80,90},
    legend pos=south east,
    ymajorgrids=true,
    grid style=dashed,
]

\addplot[
    color=black,
    mark=square,
    ]
    coordinates {
    (0.1,17.6)(0.5,70.14)(1,82.53)(3.327,85.12)};
    \addlegendentry{XLM-R$_{base}$}
    
\addplot[
    color=red,
    mark=square,
    ]
    coordinates {
    (0.1,53.29)(0.500,74.46)(1,79.98)(3.327,84.75)};
    \addlegendentry{SinBERT$_{small}$}
    
\addplot[
    color=blue,
    mark=square,
    ]
    coordinates {
    (0.100,50.42)(0.500,80.59)(1,82.32)(3.327,85.19)};
    \addlegendentry{SinBERT$_{large}$}
    
\end{axis}
\end{tikzpicture}
\caption{Results in macro-F1 score for varying dataset sizes in news category classification task with SinBERT models and XLM-R-base}
\label{fig.3}
\end{figure}
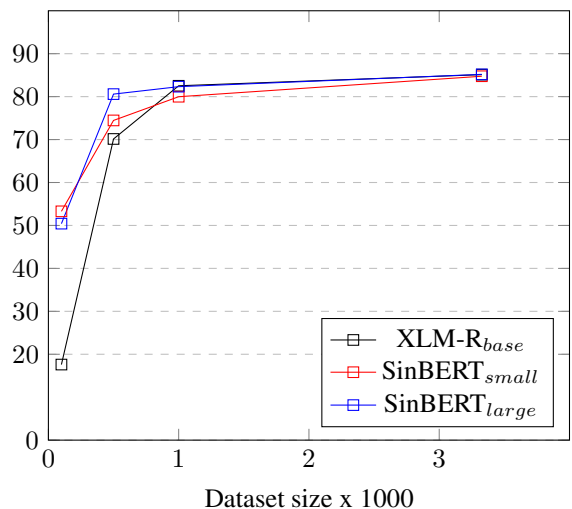

%
%
%
%
%

\begin{figure}
\begin{tikzpicture}
\begin{axis}[
    xlabel={Dataset size x 1000},
    xmin=0, xmax=13,
    ymin=0, ymax=100,
    xtick={0,2,4,6,8,10,12},
    ytick={0,10,20,30,40,50,60,70,80,90,100},
    legend pos=south east,
    ymajorgrids=true,
    grid style=dashed,
]

\addplot[
    color=black,
    mark=square,
    ]
    coordinates {
    (0.1,17.88)(0.5,84.854)(1,92.99)(10,97.018)(12.154,96.89)
    };
    \addlegendentry{XLM-R$_{base}$}
    
\addplot[
    color=red,
    mark=square,
    ]
    coordinates {
    (0.1,66.202)(0.500,89.542)(1,89.326)(10,95.164)(12.154,95.004)
    };
    \addlegendentry{SinBERT$_{small}$}
    
\addplot[
    color=blue,
    mark=square,
    ]
    coordinates {
    (0.100,76.41)(0.500,91.19)(1,91.224)(10,95.88)(12.154,95.49)
    };
    \addlegendentry{SinBERT$_{large}$}
    
\end{axis}
\end{tikzpicture}
\caption{Results in macro-F1 score for varying dataset sizes in writing style classification task with SinBERT models and XLM-R-base}
\label{fig.4}
\end{figure}
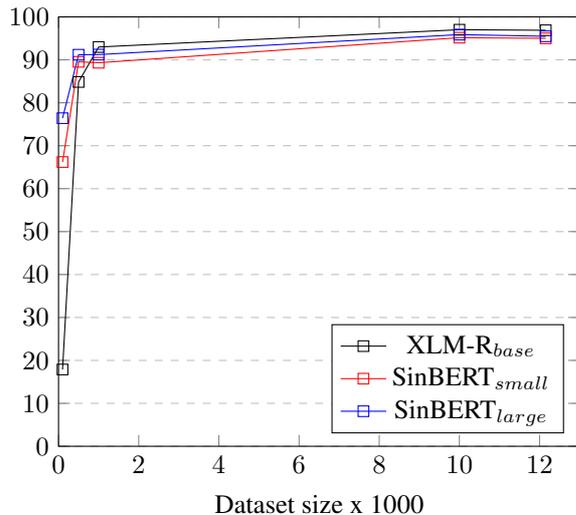

\section{Conclusion}
Although Sinhala has been included in several multilingual pre-trained language models and there exist several monolingual Sinhala pre-trained models, no empirical analysis has been conducted on their performance with respect to NLP tasks. This paper took the first step in this direction, by providing a comprehensive analysis of these models for Sinhala text classification. We also built two Sinhala pre-trained models, which have been publicly released along with the fine-tuned models. Based on the results, we provided a set of recommendations for future research that plans to use the pre-trained models for Sinhala text classification. We also showed that the XLM-R-large model sets a very strong baseline for Sinhala text classification. As an additional contribution, we release annotated datasets for Sinhala news source classification and other modified datasets (news category classification, writing style classification) that we use in our experiments. Additionally, we publicly release pre-training and fine-tuning codes. In the future, we plan to improve SinBERT with additional pre-training data and to test on more downstream tasks. 

\section{Acknowledgement}
Vinura Dhananjaya was funded by a Senate Research Committee (SRC) grant of University of Moratuwa, Sri Lanka. We also acknowledge the Prof.~V.K. Samaranayake research grant provided by LK Domain Registry for funding the conference participation.
\section{Bibliographical References}\label{reference}

\bibliographystyle{lrec2022-bib}
\bibliography{lrec2022-example}


\end{document}